\documentclass[preprint,12pt]{elsarticle}

\usepackage{amssymb}
\usepackage{amsthm}
\usepackage{mathrsfs}

\usepackage{mathtools}
\usepackage{amsfonts,amsmath,amscd}
\usepackage{xcolor}
\usepackage{subcaption,graphicx,geometry,setspace}
\usepackage{url,multirow,enumitem,float,subfloat}

\usepackage{algorithm}
\usepackage{algpseudocode}
\usepackage{lineno}

\theoremstyle{definition}
\newtheorem{definition}{Definition}

\begin{document}

\begin{frontmatter}

\title{Leveraging convergence behavior to balance conflicting tasks in multi-task learning}

\author[addr1]{Angelica Tiemi Mizuno Nakamura}
\ead{tiemi.mizuno@usp.br}
\author[addr2]{Valdir Grassi Jr}
\ead{vgrassi@usp.br}
\author[addr1]{Denis Fernando Wolf}
\ead{denis@icmc.usp.br}
\address[addr1]{Mobile Robotics Laboratory, Institute of Mathematics and Computer Sciences, University of S\~ao Paulo, S\~ao Carlos, Brazil.}

\address[addr2]{Department of Electrical and Computer Engineering, São Carlos School of Engineering, University of S\~ao Paulo, S\~ao Carlos, Brazil.}

\begin{abstract}
Multi-Task Learning is a learning paradigm that uses correlated tasks to improve performance generalization. A common way to learn multiple tasks is through the hard parameter sharing approach, in which a single architecture is used to share the same subset of parameters, creating an inductive bias between them during the training process. Due to its simplicity, potential to improve generalization, and reduce computational cost, it has gained the attention of the scientific and industrial communities. However, tasks often conflict with each other, which makes it challenging to define how the gradients of multiple tasks should be combined to allow simultaneous learning.
To address this problem, we use the idea of multi-objective optimization to propose a method that takes into account temporal behaviour of the gradients to create a dynamic bias that adjust the importance of each task during the backpropagation. The result of this method is to give more attention to the tasks that are diverging or that are not being benefited during the last iterations, allowing to ensure that the simultaneous learning is heading to the performance maximization of all tasks. As a result, we empirically show that the proposed method outperforms the state-of-art approaches on learning conflicting tasks. Unlike the adopted baselines, our method ensures that all tasks reach good generalization performances.

\end{abstract}


\end{frontmatter}

\renewcommand*{\today}{December 15, 2021}
\section{Introduction}

Machine learning can be used on several types of applications, and many of them may require many abilities. For example, self-driving cars must be able to detect obstacles for collision avoidance, to detect navigable areas and horizontal lane markings for lane keeping, detect and recognise traffic signs to respect traffic rules, and so on. Each of these abilities can be interpreted as a task to be performed at the same time. One common way of resolving multiple tasks is by implementing one solution for each task and running all them in parallel. However, this approach can be computationally very costly. Alternatively, there is a machine learning concept that allows simultaneous learning of multiple tasks, i.e., Multi-Task Learning (MTL). This approach has the potential of reducing the inference latency and computational storage when using a single architecture by sharing common knowledge among related tasks. Moreover, MTL can also be used to explore the idea of creating an inductive bias between tasks \citep{Caruana1997a}. Due to its potential to improve generalization performance, it has gained attention in several areas of scientific and industrial communities, such as computer vision \citep{Sener2018,Kendall2017,Chen2017b,Nakamura2021} and natural language processing \citep{Mao2020,Liu2016,Xiao2018}.

In deep neural networks, MTL can be categorized according to how parameters are shared between tasks, i.e., soft and hard parameter sharing approaches. In soft parameter sharing, each task has its own model with its own parameters, and the inductive bias occurs only by a regularization function between models. On the other hand, the hard parameter sharing creates the inductive bias through multiple tasks by sharing the same subset of parameters. This work focus on the second approach.

The main advantage of hard parameter sharing is related to the fact that multiple tasks are learned using a single network. It allows to implement simpler solutions, to reduce computational cost and inference latency at the same time it potentially improves generalization by implicit bias induction.
In this approach, the simultaneous learning of multiple tasks is usually performed by a linear combination of each task's gradients over the shared neurons \citep{Uhrig2016,Kendall2017}. Therefore, the tasks will share and compete for the same resource (e.g., neurons or convolutional kernels), making it necessary to handle conflicting tasks.
We define two tasks as conflicting when their gradients have different directions. Thus, when the descent direction of a task is not representative for the other, optimizing the network parameters in relation to a bad direction can harm the performance of some tasks. Therefore, it is necessary to find a trade-off between tasks, which is beyond what a linear combination can achieve.

As an alternative, \cite{Sener2018} proposed to model MTL as a multi-objective optimization problem to find a solution that is not dominated by any other. That is, to find a trade-off in which it is not possible to improve one task without degrading another. The problem is that, in multi-objective optimization, there is no single solution that is optimal for all tasks, but several solutions with different trade-offs between them \citep{Miettinen1998,Ehrgott2005}. Furthermore, considering problems of different natures, variations in the task's gradient scales can influence the definition of the solution. Thus, when a solution is said to be optimal, it does not mean that all tasks will be performing well, but only that the solution has reached a point where no further improvement can be made to all tasks simultaneously. This idea will be better explained in Section \ref{sec:method}. 

Based on the above considerations, this work proposes a method that guarantees better performances to all tasks, without giving up the optimality condition of multi-objective optimization.
The difference between the existing methods and the proposed one is the way how gradients are combined. 

The existing methods only assume the gradients computed in the current iteration to estimate an appropriate combination of them in order to obtain a common descent direction between tasks. However, since these methods get only a snapshot of the current gradients, they need to create a static bias when combining the gradients, such as getting the direction with the minimum norm \citep{Sener2018} or the central direction between gradients \citep{katrutsa2020}. Because of this, they cannot ensure that all tasks are converging to their optimal performances, although they are meeting the optimality conditions. On the other hand, if the temporal factor of the gradients is considered to compute their combination on the current iteration, it would be possible to analyze whether the performances of all tasks are converging or not and create a dynamic bias. For this, the proposed method creates the idea of tensors that strengthen when their respective tasks lose performance and weaken when they increase performance. This is done at the same time the optimality conditions are met. By doing so, the combination of gradients is modulated to give more preference to those that are diverging from its optimal performance, ensuring that all tasks will converge together. 

Therefore, the main contributions proposed in this paper are as follows: 1) A method to create a dynamic learning bias by leveraging the convergence behavior of tasks. 2) An approach that is independent of a specific application to combine multiple gradient vectors and find a common feasible descent direction, aiming to maximize the performance of all tasks. To confirm the performance of the proposed method in learning conflicting tasks, we performed ablation studies and a series of experiments on a public handwritten digit classification dataset.

The remainder of this paper is structured as follows. Section \ref{sec:relatedWork} provides a literature review. In Section \ref{sec:method}, we present our proposed method to combine gradients of multiple tasks by leveraging their convergence behavior over the past iterations. In Section \ref{sec:experiments}, we perform ablation studies and compare our proposed method with other weighting methods, followed by a discussion of the results. Finally, we conclude in Section \ref{sec:conclusion}.  
\section{Related Work} \label{sec:relatedWork}

Introduced by \cite{Caruana1993}, the Multi-Task Learning (MTL) is a machine learning paradigm in which multiple related tasks are learned simultaneously, such that each task can contribute to a shared knowledge that enhances the generalization performance of every other tasks at hand. This idea of learning multiple tasks may cause confusion among researchers. For example, one could say that MTL applied to neural networks is simply the fact of creating an architecture with multiple outputs, which has already been extensively explored. However, this kind of solution usually encodes a single task \citep{LeCun1989, Caruana1997a,Zhang2021a}. On the other hand, the MTL aims to explore the idea that related tasks can create an inductive bias, and based on that bias, a certain task will prefer a hypothesis $x$ out of the $n$ possible hypothesis. In deep neural networks, multi-task learning can be categorized according to how parameters are shared between tasks, i.e., soft or hard parameter sharing.

In soft parameter sharing, each task has its own model with its own parameters, and the bias induction is accomplished through functions that reduce the distance between the models' parameters \citep{Vandenhende2019, Misra2016, Ruder2017b,Liu2018b}. However, this approach results in very complex architectures with higher computational cost, which makes it unfeasible for many applications. On the other hand, the hard parameter sharing approach allows a set of parameters to be shared among all tasks, reducing inference latency and storage costs \citep{Caruana1993}. In computer vision, the architecture of a hard parameter sharing models is usually composed of a shared encoder and multiple task-specific decoders, in which the models are trained by optimizing the gradient combination of all tasks. This is commonly done by a linear combination of the negative gradients of each task \citep{Uhrig2016,Chennupati2019,Teichmann2018,Sanchez2019}, or some variation with adaptive weights \citep{Kendall2017,Chen2017b,Liu2019,Li2016a,Guo2018}. 

The main challenge of this approach is to define how the gradients of different tasks should be combined. Since the tasks share the same parameters and compete for the same resources (e.g. convolutional kernels from encoder), this combination must be done by finding a trade-off between them so that shared layers can learn relevant features to all tasks \citep{Kendall2017,Nakamura2021}. However, the linear combination does not allow finding this balance between tasks \citep{Sener2018,Lin2019}. As an alternative, \cite{Sener2018} proposed to model multitasking learning as a multi-objective optimization problem to find a solution that is not dominated by any other (Pareto optimal solutions), achieving better results than previous methods of linear combination and its variations. However, in problems with variations in the gradients' scale, these variations can significantly impact the final solution. In \cite{Sener2018}, his solution may end up prioritizing only the tasks that have a lower gradient norm and not learning the other tasks well. 
To find a solution that considers the tasks equally important, \cite{Harada2006,katrutsa2020} remove the scale influence by normalizing the gradients for each task. In \cite{Mao2020}, it is proposed to use the Tchebycheff metric to handle non-convex problems. Instead of finding a single optimal solution, \cite{Lin2019,Mahapatra2020} propose to find a set of Pareto optimal solutions with different trade-offs between tasks. 

However, these methods consider the current tasks' gradients to find a common descent direction, not analyzing the previously defined descent directions or any indication if they are being representative to learn all tasks.
It creates a tendency for the solution to prefer a specific direction. Therefore, this work proposes to analyze the convergence of tasks during the training process to dynamically correct the common descent direction and find a trade-off between conflicting tasks that maximizes simultaneous learning. Thus, we define a new descent direction that prevents the model from learning more about a specific task by pulling tasks that are diverging or that are not being benefited during training. Through experimental results, we show that the proposed method can handle conflicting tasks, finding a good balance between them during the training process.

\section{Approach}\label{sec:method}

\begin{figure}[th]
    \centering
    \begin{subfigure}{.45\textwidth}
        \centering
        \includegraphics[width=\textwidth]{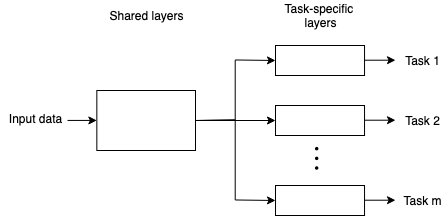}
        \caption{Feedforward}
    \end{subfigure}
    \hspace{0.00mm}
    \begin{subfigure}{.45\textwidth}
        \centering
        \includegraphics[width=\textwidth]{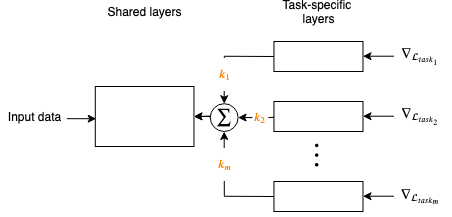}
        \caption{Backpropagation}
    \end{subfigure}
    \caption{Multi-task learning architecture. The output from the shared layers is split into a set of $m$ task-specific layers. During backpropagation, the parameters from task-specific layers are optimized according to the gradients of each task, while in the shared layers, the optimization occurs by combining the gradients of all tasks. This combination is usually performed by a weighted sum, in which each task is associated with a weighting coefficient, $k_m$.}
    \label{fig:fowback}
\end{figure}

The idea of Multi-task Learning (MTL) is that related tasks can create an inductive bias, and based on that bias, a certain task will prefer some hypothesis over the $n$ possible hypotheses \citep{Caruana1993}. To perform simultaneous learning of multiple tasks, we will consider a hard parameter sharing approach,  in which a subset of hidden layers parameters with low-level features are shared between all tasks, while the rest of the high-level parameters are kept as task-specific (Figure \ref{fig:fowback}). Hence, considering a multi-task learning problem over a space $X$ and a set of tasks  $ \{Y_m \}_{m \in [M]} $, the decision function of each task $m$ can be defined as $f_m(x;\boldsymbol\theta_{sh};\boldsymbol\theta_m): X \rightarrow Y$. Where the  $\boldsymbol\theta_{sh}$ parameters are shared among tasks and  $\boldsymbol\theta_{m}$ are task-specific parameters. 

Given a set of objective functions $\mathscr{L}(f_m(x;\boldsymbol{\theta}_{sh},\boldsymbol{\theta}_m)) $ that defines the empirical risk of the task $m$, the MTL problem is commonly formulated as the minimization of the linear combination of objective functions:

\begin{equation*}
\min_{\scriptstyle \boldsymbol{\theta}_{sh}, \atop\scriptstyle \boldsymbol{\theta}_1, \ldots,\boldsymbol{\theta}_M}  \sum_{m=1}^{M}\frac{1}{N}\sum_{i=1}^{N} k_m \mathscr{L}(f_m(x_i;\boldsymbol{\theta}_{sh},\boldsymbol{\theta}_m), y_{i,m}),
\label{Eq:MTLloss}
\end{equation*}

\noindent where $y_{i,m}$ is the label of the $m^{th}$ task of the $i^{th}$ data, $N$ is the number of data and $k_m>0$ are the predefined or dynamically calculated weighting coefficients associated with each objective function. 

Although the linear combination is a simple method to solve multi-task learning problem, this approach cannot adequately model competing tasks \citep{Sener2018,Mahapatra2020}. On the other hand, multi-objective optimization is an optimization problem that deals with multiple conflicting objective functions, which makes it an intuitive way to deal with problems where there is no single optimal solution for all objective functions \citep{Miettinen1998,Ehrgott2005}.

\subsection{Feasible descent direction}

\begin{figure}[h]
    \centering
    \includegraphics[width=10cm]{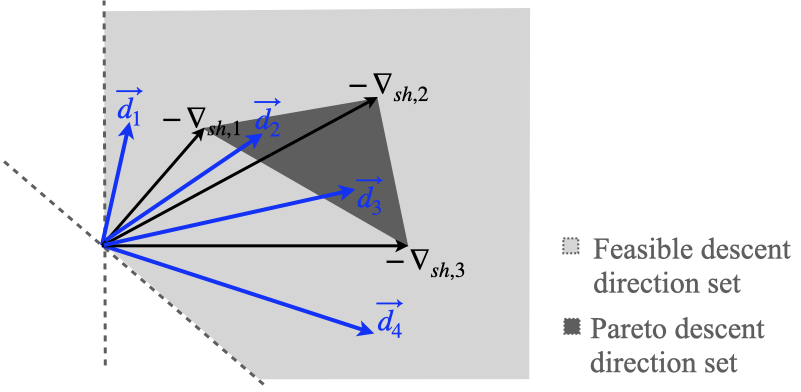}
    \caption{Feasible set and Pareto descent directions considering the negative gradient vectors of $\nabla_{sh,1}$, $\nabla_{sh,2}$ and $\nabla_{sh,3}$. The blue vectors are examples of feasible descent directions given the set of negative gradients of the tasks. Best visualized with color.}
    \label{fig:feasible}
\end{figure}

According to \cite{Fliege2000}, a direction $\Vec{v}$ can simultaneously improve all tasks if:

\begin{equation}
    -\nabla_{sh,m}\cdot \Vec{v} \geq 0, \forall m \in {1,\dots, M},
    \label{eq:feasible_dir}
\end{equation}

\noindent where $\nabla_{sh,m}$ represents the gradient vector of the $m^{th}$ task w.r.t to the shared parameters, $sh$.
Thus, given the negative gradient vectors of three tasks, there will be several feasible descent directions. 
The set of all feasible descent directions are shown in gray in Figure \ref{fig:feasible}.

However, not all feasible descent directions will be representative for all objective functions. For example, considering the feasible descent directions $\Vec{d_1}$, $\Vec{d_2}$, $\Vec{d_3}$ and $\Vec{d_4}$, we can see that $\Vec{d_2}$ decreases more each objective function than $\Vec{d_1}$:
\begin{linenomath}
\begin{align*}
\Vec{d_1} \cdot -\nabla_{sh,1}(\boldsymbol{\theta}) &< \Vec{d_2} \cdot -\nabla_{sh,1}(\boldsymbol{\theta}),  \\
\Vec{d_1} \cdot -\nabla_{sh,2}(\boldsymbol{\theta}) &< \Vec{d_2} \cdot -\nabla_{sh,2}(\boldsymbol{\theta}), \\
\Vec{d_1} \cdot -\nabla_{sh,3}(\boldsymbol{\theta}) &< \Vec{d_2} \cdot -\nabla_{sh,3}(\boldsymbol{\theta}).
\end{align*}
\end{linenomath}

Similarly, the same holds for $\Vec{d_3}$ and $\Vec{d_4}$. On the other hand, $\Vec{d_2}$ and $\Vec{d_3}$ descent directions cannot improve an objective function without degrading the improvement of another:
\begin{linenomath}
\begin{align*}
\Vec{d_2} \cdot -\nabla_{sh,1}(\boldsymbol{\theta}) &> \Vec{d_3} \cdot -\nabla_{sh,1}(\boldsymbol{\theta}),  \\
\Vec{d_2} \cdot -\nabla_{sh,2}(\boldsymbol{\theta}) &> \Vec{d_3} \cdot -\nabla_{sh,2}(\boldsymbol{\theta}),\\ 
\Vec{d_2} \cdot -\nabla_{sh,3}(\boldsymbol{\theta}) &< \Vec{d_3} \cdot -\nabla_{sh,3}(\boldsymbol{\theta}),
\end{align*}
\end{linenomath}

\noindent being necessary to find a trade-off between the objectives. These solutions are known as Pareto optimal solutions (dark gray region in Figure \ref{fig:feasible}).

\subsection{Pareto descent directions}

\begin{definition}[Pareto-optimality]  
A decision vector $\mathbf{x^*} \in S$ is Pareto optimal if there does not exist another decision vector $\mathbf{x}\in S$ such that:
\begin{enumerate}
    \item $f_i(\textbf{x}) \leq f_i(\textbf{x}^{*}), \forall i \in \{1,\dots, M\}$;
    \item $f_j(\textbf{x}) < f_j(\textbf{x}^{*}), \text{ for some } j \in \{1,\dots, M\} $,
\end{enumerate}

\end{definition}

A Pareto optimal solution can be expressed as a convex combination of the negative gradients \citep{Harada2006,Miettinen1998}. In \cite{Desideri2012}, it was introduced the notion of Pareto-stationarity as a necessary condition for Pareto-optimality: 

\begin{definition}[Pareto-stationarity] 
Given M differentiable objective functions $\mathscr{L}_m(\boldsymbol{\theta}), \\m=1, \cdots, M, \; \boldsymbol{\theta} \in {\rm I\!R^d}$, a point $\boldsymbol{\hat{\theta}}$ is considered a Pareto-stationary point, if and only if there exists a convex combination of the gradients that is equal to zero, i.e.:
\begin{center}
    $\sum_{i=1}^M k_m \nabla_{sh,m}(\boldsymbol{\hat{\theta}}) = 0$, \; where $\sum_{i=1}^M k_m = 1, k_m \geq 0 \quad \forall m$.
\end{center}
\label{def:pareto}
\end{definition}
 Therefore, if a given point $\boldsymbol{\hat{\theta}}$ is not Pareto-stationary, it will be a descent direction for all objective functions \citep{Schaffler2002,Desideri2012}.
 
 Based on Definition \ref{def:pareto}, a common descent direction can be defined by searching for weighting coefficients that minimize the norm of the convex combination \citep{Sener2018,Desideri2012}:
 
 \begin{equation}
    \operatorname*{argmin}_{k_1, \cdots, k_M}   \bigg\{ \Big\|\sum_{m=1}^M k_m \nabla_{sh,m}\Big \|_2^2 \bigg| \sum_{m=1}^M k_m=1, \, k_m \geq 0 \quad \forall m \bigg\}.
    \label{eq:sener} 
\end{equation}

\begin{figure}[h]
    \centering
    \includegraphics[width=12cm]{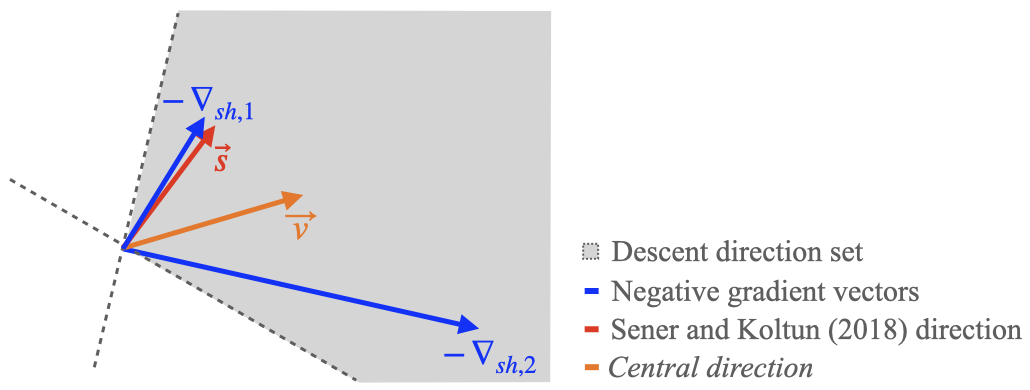}
    \caption{Common descent directions considering the negative gradient vectors of $\nabla_{sh,1}$ and $\nabla_{sh,2}$. Best visualized with color.}
    \label{fig:feasible_dir}
\end{figure}

This solution will result in the minimal-norm element in the convex hull of the negative gradients. However, when we deal with problems of different natures in multi-task learning, we may also have to deal with variations in the order of gradients' magnitude. Thus, when a task is given more importance then another during training, the final model's performance can be harmed. Although the descent direction found by Equation \ref{eq:sener} is based on an optimality condition (Definition 1), this solution may not be representative for all tasks when the tasks' losses are not balanced. In this case, the solution will tend to give more preference to the direction with least norm (red vector in Figure \ref{fig:feasible_dir}). 

To avoid prioritizing only the direction of a specific task, we remove the scale influence on the problem by considering the unit vector of each task's gradient, $\Bar{\nabla} _{sh,m} = \nabla_{sh,m}/ \| \nabla_{sh,m} \|$. Therefore, all tasks will be equally important during the descent direction computation. The Equation \ref{eq:sener} can be rewritten as:

\begin{equation}
    \operatorname*{argmin}_{k_1, \cdots, k_M}   \bigg\{ \Big\|\sum_{m=1}^M k_m \bar{\nabla}_{sh,m}\Big \|_2^2 \bigg| \sum_{m=1}^M k_m=1, \, k_m \geq 0 \quad \forall m \bigg\}.
    \label{eq:sener_norm} 
\end{equation}

Solving optimization problem (\ref{eq:sener_norm}), the common descent direction can be obtained by a convex combination of each task's gradients with their respective weighting coefficients, $\sum_{m=1}^M k_m\bar{\nabla}_{sh,m}$.
This direction will be close to the central region of the cone formed by the negative gradient vectors of different tasks 
(orange vector in Figure \ref{fig:feasible_dir}), which will result in the same relative decrease among them. 
Since this direction was calculated considering unit vectors, we must recover the scale factor so that the resulting vector lies in the convex hull formed by the negative gradient vectors of the tasks. This can be done using a scale factor $\gamma$ as proposed in \cite{katrutsa2020}. Therefore, the resulting common descent direction, $\Vec{v}$, that lies in the central region between tasks can be defined as:

\begin{equation}
    \Vec{v} = \gamma \sum_{m=1}^{M}k_m\bar{\nabla}_{sh,m},
    \label{eq:central_dir}
\end{equation}

\noindent where $\gamma = \Big ( \sum_{m=1}^{M} \frac{k_m}{\left \| \nabla_{sh,m} \right \|}\Big )^{-1}$.

However, note that this direction and all others studied so far create an assumption that the direction found is representative considering only the gradient analysis at a current time, without considering the history or any indication that the directions defined during the training process were being representative for all tasks. This means that, regardless of the method adopted to compute a common descent direction, the direction found will tend to prefer a specific descent direction (i.e.,  direction that results in lesser or greater magnitude, central direction, etc.). Thus, to enable the correction of the common descent direction 
during training, we propose to analyse the convergence of the models by tracking the gradients' norm of each task during a period of $T$ iterations and adopt the idea of tensioners that pull the common descent direction for the task that is diverging. 

\begin{figure}[h]
    \centering 
    \begin{subfigure}{0.48\textwidth}
        \includegraphics[width=\textwidth]{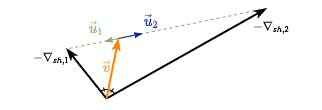}
        \caption{}
    \end{subfigure}
    \begin{subfigure}{0.48\textwidth}
        \includegraphics[width=\textwidth]{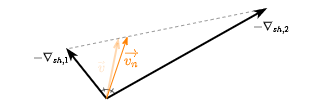}
        \caption{}
    \end{subfigure}
    \caption{Method for changing the common descent direction between tasks. (a) Tensioners that will pull the common descent direction for the diverging task. (b) New defined descent direction. }
   \label{fig:change}
\end{figure}

\subsection{Task tensioner}

Given the central direction, $\Vec{v}$, obtained through the Equation \ref{eq:central_dir}, the tensions of each task can be computed according to the following equation:

\begin{equation}
    u_m = c_m\Bigg(\frac{\nabla_{sh,m} - \Vec{v}}{\|\nabla_{sh,m} - \Vec{v}\|}\Bigg), \quad \forall m=1,\dots, M,
    \label{eq:tensors}
\end{equation}

\noindent where $\|\cdot\|$ is l2-norm, and the $c_m$ factor measure how much each task must be pulled to correct the common descent direction considering a relative variation of gradients' norm during a predefined number of iterations (Figure \ref{fig:change}). Therefore, given the gradient vector of each task, the average of gradients' magnitude accumulated during a period of T iterations is  calculated by:  
\begin{linenomath}
\begin{equation*}
    \zeta_m(t)= \frac{1}{T}\sum_{i=0}^{T-1}\left \| \nabla_{sh,m}(t-i) \right \|.
\end{equation*}
\end{linenomath}
Then, the relative variation of the gradient's norm can be obtained by the ratio between current and previous accumulated gradient:

\begin{equation}
    \delta_m =  \frac{\left\|\zeta_m(t)\right\|}{\left\|\zeta_m(t-1)\right\|} + \log_{10}(\mathscr{L}_m).
    \label{eq:relat_cange}
\end{equation}

The term $\log_{10}(\mathscr{L}_m)$ aims to penalize tasks that have higher loss, increasing the tension factor of the respective tasks.
Therefore, given the value of $\delta_m$, we can get the tension factor for each task $m$:

\begin{equation}
    c_ m = \frac{\alpha}{1+e^{(-\delta_m\cdot e+e)}}+1-\alpha.
    \label{eq:tens_factor}
\end{equation}

As we want to change the descent direction only if some task is diverging, the  $c_m$ values are calculated so that the direction does not change too much from the convergence region of all tasks. Thus, a constant variable, $\alpha \in [0,1]$, is used to regulate the  tension factor's sensitivity. Therefore, if $\alpha=0$, the resulting vector will be exactly the vector $\Vec{v}$. And, the higher the alpha value, the more the common descent direction will be pulled towards the task that is under-performing or diverging.

The new common descent direction among tasks can be defined by the sum of the current common descent direction, $\Vec{v}$, with the linear combination of each task's tension (Equation \ref{eq:tensors}):

\begin{equation}
    \Vec{v}_n = \Vec{v} + \sum_{m=1}^M c_m\Bigg(\frac{\nabla_{sh,m} - \Vec{v}}{\|\nabla_{sh,m} - \Vec{v}\|}\Bigg),
    \label{eq:new_dir}
\end{equation}

\noindent subject to $\nabla_{sh,m} \cdot \Vec{v}_n \geq 0, \forall m = 1, \dots, M$.

Algorithm \ref{alg:grad_update} shows the proposed method that takes into account gradient history during training to pull the common descent direction for the diverging task.

\begin{algorithm}[h]
\caption{Gradient accumulation}\label{alg:cap}
\begin{algorithmic}[1]
\For {$i = {0, ..., I-1}$}
    \For {$m={1,\cdots}, M$}
        \State $\mathscr{L}_m \gets \mathscr{L}(f_m(x_i;\theta_{sh}, \theta_m), y_{i,m})$ \Comment{Compute task-specific loss}
        \State  $\nabla_m \gets \frac{\partial\mathscr{L}_m}{\partial \theta_m}$ \Comment{Gradient descent on task-specific parameters}
        \State $\nabla_{sh,m} \gets \frac{\partial\mathscr{L}_{m}}{\partial \theta_{sh}}$ \Comment{Gradient descent on shared parameters}
    \EndFor
    \State Solve (\ref{eq:sener_norm}) to obtain $k_m$ and find common descent direction $\sum k_m\bar{\nabla}_{sh,m}$
    \If{$i \geq T$}
        \State Compute new common descent direction using Algorithm 2.
    \Else 
        \State $\Vec{v}_n \gets \Vec{v}$
    \EndIf
    \State Update $\theta_{sh}$ w.r.t $\Vec{v}_n$ with chosen optimizer;
    \State Update $\theta_{m}, \; \forall m$ with chosen optimizer.
\EndFor
\end{algorithmic}
\label{alg:grad_update}
\end{algorithm}

\begin{algorithm}[h]
\caption{Compute task tension}\label{alg:cap}
\begin{algorithmic}[1]
\Require $i \geq T$
    \For {$m={1,\cdots}, M$}
    \State Compute relative change, $\delta_m$  (\ref{eq:relat_cange})
    \State Compute tension factor, $c_m$ (\ref{eq:tens_factor})
    \State Compute task tension, $u_m$ (\ref{eq:tensors})
    \EndFor
    
\State $\Vec{v}_n \gets \Vec{v}+\sum_{m=1}^M u_m$
\Comment{Update common descent direction (\ref{eq:new_dir})}
\State \Return $\Vec{v}_n$
\end{algorithmic}
\end{algorithm}

\section{Experiments}\label{sec:experiments}

To analyze the behavior of different descent direction's methods, and ensure that it is not influenced by the network architecture or other variables, ablation studies are performed considering a Multi-layer Perceptron (MLP) to learn problems of multiple logical operators. Afterward, the evaluation is performed in an image domain application considering the classification and reconstruction problems in the MNIST dataset adapted to perform multi-task learning.

\subsection{Experimental setup and metrics}
The experiments were conducted using a machine with Intel Core i7-10750H, 16GB RAM, and RTX 2060 6GB. For all experiments described in this section, it was considered the values of  $\alpha=0.3$ (Equation \ref{eq:tens_factor}) and period ${T}={10}$ (Equation \ref{eq:relat_cange}).

For quantitative evaluations, we used the mean squared error (MSE) and classification accuracy:

\begin{align*}
MSE &= \frac{1}{n}\sum_{i=1}^{n}\big(y_i-\hat{y}_i\big)^2,  \\
Accuracy &= \frac{TP+TN}{TP+FP+FN+TN}, 
\end{align*}

\noindent where $\hat{y}$ is the estimated value, $y$ is the ground truth, $n$ is the total number of pixels in the evaluated images. TP, TN, FP, and FN are the numbers of true positive, true negative, false positive, and false negative classifications, respectively.

\subsection{Ablation Studies}

\begin{figure}[h]
    \centering 
    \begin{subfigure}{0.4\textwidth}
        \includegraphics[width=\textwidth]{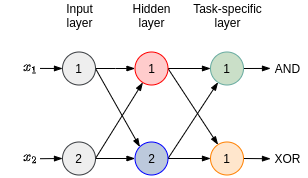}
        \caption{}
        \label{fig:mlp_1}
    \end{subfigure}
    \begin{subfigure}{0.58\textwidth}
        \includegraphics[width=\textwidth]{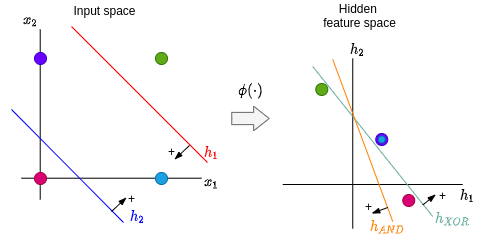}
        \caption{}
        \label{fig:mlp_2}
    \end{subfigure}
     
    \caption{(a) MLP for AND and XOR tasks. (b) Feature space and hyper-planes necessary to separate the input data. Best visualized with color.}
   \label{fig:mlp}
\end{figure}

In these ablation studies, the simultaneous learning of the logical operators XOR and AND was evaluated.
We adopted this study because we know exactly how many neurons are needed to generate the hyperplanes that separate the sets of tasks' data in the feature space. 
Thus, we eliminate external influences that generate uncertainty if it was not possible to obtain a better result due to the architecture or problem modeling, allowing to evaluate only the descent direction methods. Furthermore, the problems are both complementary and conflicting. Since in shared neurons, the AND task will try to separate the feature space into two regions, while the XOR task will need to separate the feature space into three regions. Figure \ref{fig:mlp_1} shows a MLP with two shared neurons and two task-specific output neurons. The hyper-planes necessary to separate the data of both tasks are shown in Figure \ref{fig:mlp_2}.

The experiments were performed considering the common descent directions computed according to the methods of uniform weighting, minimum-norm \citep{Sener2018}, central direction \citep{katrutsa2020}, and  gradient modification with tasks' tensors (proposed). 
To analyse the descent methods considering tasks with different orders of magnitude, the annotations of XOR task were multiplied by a constant $c\geq1$  to change the problem's scale. 

Each model was trained using mean squared error (MSE) loss function and stop condition with the limit of 3,000 epochs or an error smaller than $1\mathrm{e}{-3}$. 
Since training results can vary from run to run, each method was trained 50 times with a different seed value to get deterministic random data. Furthermore, the models were trained considering a set $lr=\{5\mathrm{e}{-5},1\mathrm{e}{-4},5\mathrm{e}{-3},1\mathrm{e}{-3},5\mathrm{e}{-2},1\mathrm{e}{-2}\}$ of learning rates and the Stochastic Gradient Descent (SGD) optimization algorithm with momentum. Thus, the reported results are the models that best separated the input data. 
\newline\newline
\noindent\textit{\textbf{Similar gradients' magnitude:}} Table \ref{tab:mlp_xor1} presents the average performance of each method considering $c=1$, i.e., without much difference in the gradients' magnitude. The first two rows correspond to the models separately trained for each task, which we define as our baseline. 

As can be seen, all methods were able to create hyperplanes capable of separating the data for both tasks and without significant difference in the convergence time. With the exception of the method proposed by \cite{Sener2018}, which has a slightly slower convergence compared to other methods because it prioritizes the task direction with a lower gradient magnitude.

Regarding the use of the tensors proposed in the present work to change the common descent direction according to each task's learning progress, the performance must be similar or better than the unaltered directions. As can be seen, the proposed method resulted in a faster convergence. In the case of the central direction,  the use of tensors reduced the number of epochs needed from 574.44 to 443.02 epochs, resulting in the method that obtained the fastest convergence. And when we apply the tensors in the Sener method, which was the method that had the slowest convergence, it was possible to reduce from 758.56 to 493.16 epochs.

\begin{table}[H]
\centering
\begin{tabular}{rcc}
\hline
\multicolumn{1}{c}{\begin{tabular}[c]{@{}c@{}}Descent direction \\ method\end{tabular}} & \begin{tabular}[c]{@{}c@{}}Convergence \\ rate (\%)\end{tabular} & \begin{tabular}[c]{@{}c@{}}Training epochs \\ average\end{tabular} \\ \hline
Single AND                                                                              & 100.00                                                           & $351.70\pm2.62$                                                             \\
Single XOR                                                                              & 100.00                                                           & $871.20\pm52.95$                                                             \\
Uniform weighting                                                                              & 100.00                                                           & $484.86\pm18.12$                                                             \\
Min-norm                                                                                   & 100.00                                                           & $758.56\pm137.15$                                                             \\
Central dir.                                                                                & 100.00                                                           & $574.44\pm43.37$                                                             \\
Min-norm + tensor                                                                          & 100.00                                                           & $493.16\pm27.94$                                                             \\
Central dir. + tensor                                                                       & 100.00                                                           & $\mathbf{443.02\pm21.14}$ \\ \hline                   
\end{tabular}
\caption{Average performance of evaluated methods without changing the problems' scale.}
\label{tab:mlp_xor1}
\end{table}

\noindent\textit{\textbf{Different gradient's magnitude:}}
To analyze the behavior of descent direction methods considering problems closer to which the proposed method seeks to solve, i.e., when we have tasks with variation in the  gradients' magnitude due to the tasks having different types of nature or even loss functions, the annotations of the XOR task were multiplied by the constant $c=10$ to simulate this variation. 

\begin{figure}[h]
    \centering
    \begin{subfigure}{0.325\textwidth}
    \includegraphics[width=\textwidth]{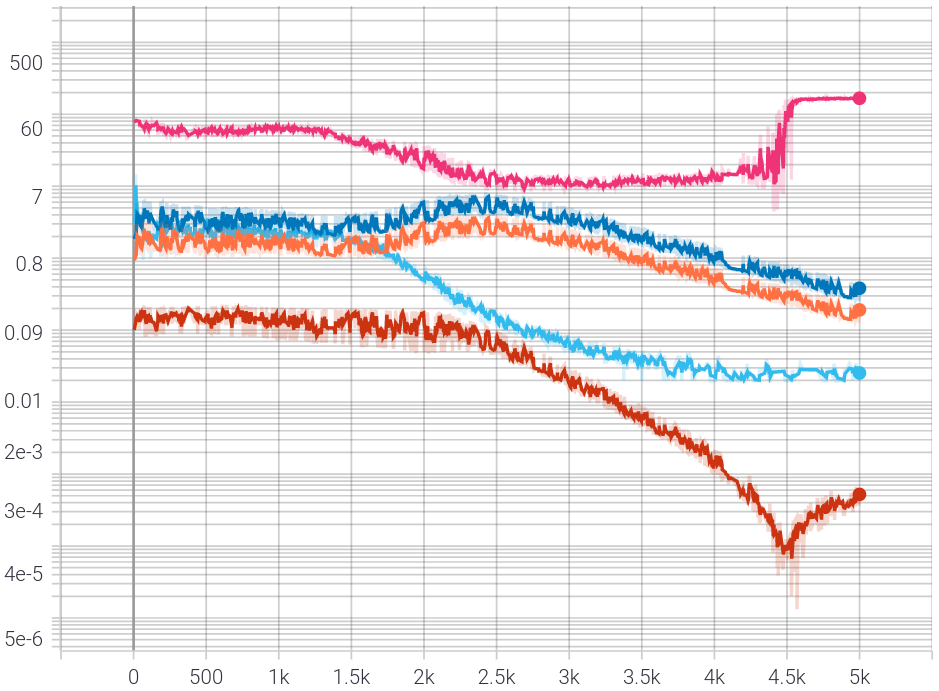}    
    \caption{Uniform}
    \label{fig:sub_uniform}
    \end{subfigure}
    \begin{subfigure}{0.325\textwidth}
    \includegraphics[width=\textwidth]{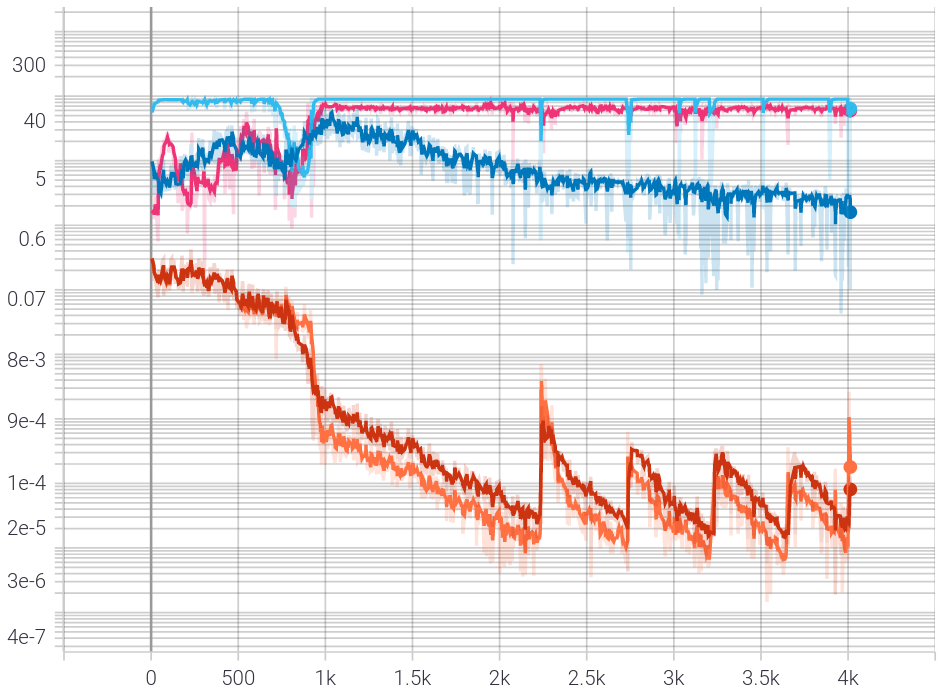}    
    \caption{Minimum-norm}
    \label{fig:sub_bisec}
    \end{subfigure}
    \begin{subfigure}{0.325\textwidth}
    \includegraphics[width=\textwidth]{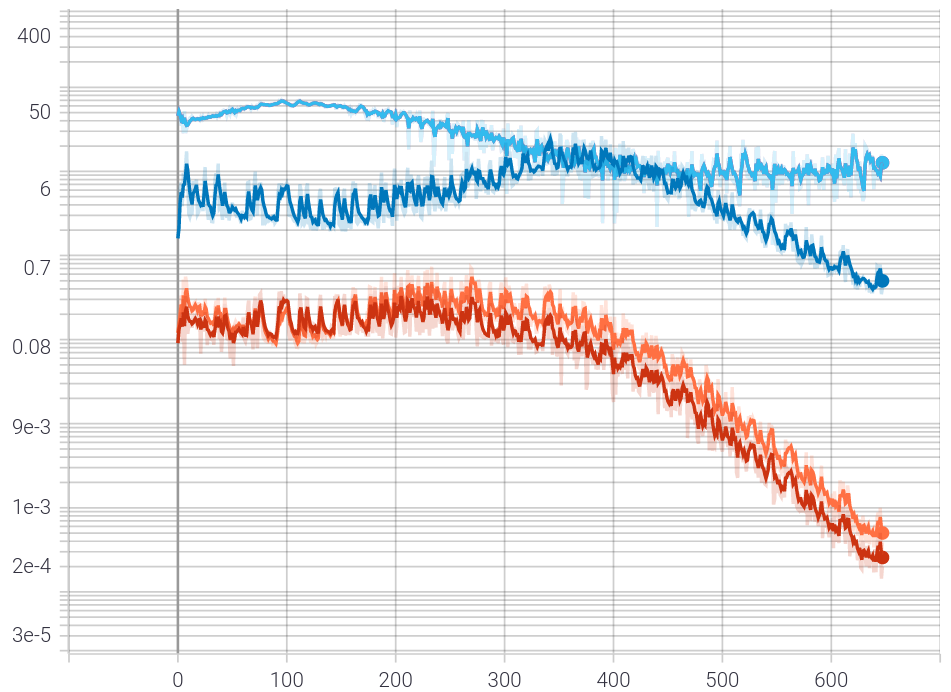}    
    \caption{Central direction}
    \label{fig:sub_sener}
    \end{subfigure}
    \begin{subfigure}{0.325\textwidth}
    \includegraphics[width=\textwidth]{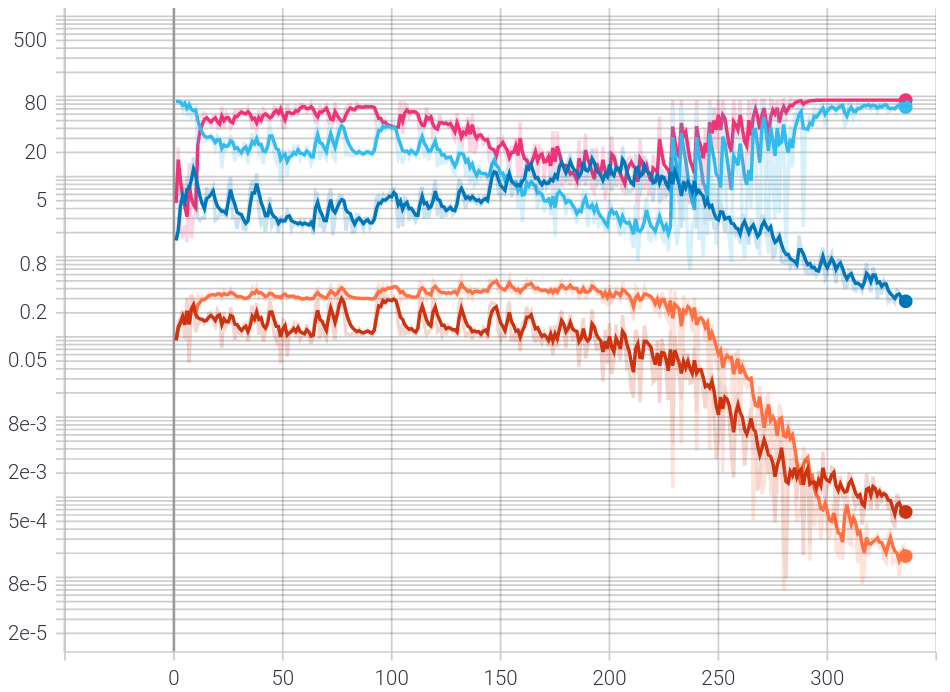}    
    \caption{Minimum-norm + tensor}
    \label{fig:sub_senerT}
    \end{subfigure}
    \begin{subfigure}{0.325\textwidth}
    \includegraphics[width=\textwidth]{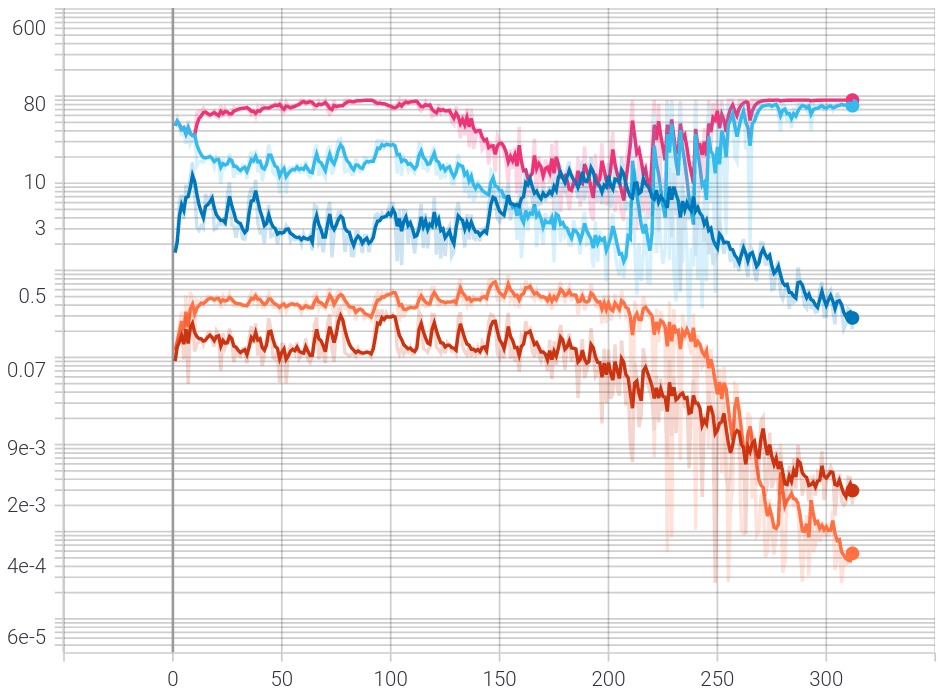}    
    \caption{Central direction + tensor}
    \label{fig:sub_bisecT}
    \end{subfigure}
    \caption{Learning curves of different descent direction methods. Pink and light blue curves represent the angles of AND and XOR tasks in relation to the common descent direction, respectively. Red and dark blue curves represents the gradients' norm of AND and XOR tasks, respectively. Finally, orange curves represents the gradient's norm of the common descent direction. Best visualized with color.}
    \label{fig:learning_curves}
\end{figure}

The behaviour of each descent direction method can be observed in the learning curves of Figure \ref{fig:learning_curves}. When uniform weighting is used, we can see that the common direction gives preference to the task with greater magnitude, in which the angle between the common descent direction and the XOR task is close to zero (light blue curve in the Figure \ref{fig:sub_uniform}). This can make the simultaneous learning to be substantially accomplished in relation to a single task, and the common descent direction does not be feasible for the other tasks. As in the case of the AND task, where in some iterations the angle between its direction and the defined common descent direction is greater than 90 degrees (Definition \ref{eq:feasible_dir}). On the other hand, when we have tasks with difference in the gradient scale, the method proposed by \cite{Sener2018} gives preference to the gradient direction with the lowest norm, making the model converge faster in relation to this task, which can hinder the learning of the other tasks or slow down the learning process (Figure \ref{fig:sub_sener}). 

The central direction is the direction close to the central region of the cone formed by the negative gradients, resulting in the same angle between tasks (light blue and pink curves are overlapping in Figure \ref{fig:sub_bisec}). Therefore, this direction tends to give equal importance to all tasks, regardless of the gradients norm. Figures \ref{fig:sub_bisecT} and \ref{fig:sub_senerT} show the minimum-norm and central directions changed with the use of tensors, respectively. 
As it can be seen, in the first iterations the central direction method assigns the same importance to both tasks, resulting in the common descent direction having the same angle between the tasks. However, we can observe that, in addition to the XOR task norm being higher, it increases significantly compared to another task (dark blue curve in the Figure \ref{fig:sub_bisecT}). From $T=10$ iterations, the use of tensors starts to dynamically change the common descent direction in relation to the task that is diverging. Thus, as soon as the norm of both tasks starts to reduce simultaneously, the common descent direction starts to give practically the same importance to the tasks (after approximately 230 iterations), with a small variation due to the penalty factor in Equation \ref{eq:relat_cange}. The same behavior can be observed when we apply the use of tensors in the direction of the minimum-norm method, proving that the proposed method can adapt to different descent directions.

\begin{table}[h]
\centering
\begin{tabular}{lcc}
\hline
\multicolumn{1}{c}{\begin{tabular}[c]{@{}c@{}}Descent direction \\ method\end{tabular}}   & \begin{tabular}[c]{@{}c@{}}Convergence \\ rate (\%)\end{tabular} & \begin{tabular}[c]{@{}c@{}}Training epochs \\ average\end{tabular} \\ \hline
Single-task AND        & 100.00                                                           & $351.70\pm2.62$                                                             \\
Single-task XOR        & 100.00                                                            & $2{,}979.00\pm0.80$                                                            \\
Uniform weighting          & 46.00                                                            & $1{,}831.60\pm810.81$                                                            \\
Min-norm        & 100.00                                                           & $1{,}998.44\pm316.38$                                                            \\
Central dir.           & 100.00                                                           & $328.50\pm25.45$                                                             \\
Min-norm + tensor    & 100.00                                                           & $210.38\pm14.96$                                                             \\
Central dir. + tensor & 100.00                                                           & $\mathbf{188.20\pm24.29}$    \\ \hline

\end{tabular}
\caption{Average performance of evaluated methods considering $c=10$ for XOR task.}
\label{tab:mlp_xor10}
\end{table}

The average of the $50$ trainings are shown in the Table \ref{tab:mlp_xor10}. Since only the scale of the XOR problem was changed, the performance of the model trained separately for AND task remains the same. In this experiment, the impact of using different descent directions' methods becomes more visible. For the single-task XOR, the model needs more iterations to converge when the scale's problem changes. When performed the simultaneous learning of both tasks, the uniform weighting resulted in the worst performance, being able to separate the feature space only in $46\%$ of the $50$ trainings, and, in the best case, needing more than $1,000$ epochs for convergence.
Regarding the methods of central direction and minimum-norm \citep{Sener2018,katrutsa2020}, both converged in all trainings. However, as can be seen in the fourth row of Table \ref{tab:mlp_xor10}, the method proposed by \cite{Sener2018} needs more iterations. Finally, when we use the tensors to change the central and the minimum-norm direction, we again improve the convergence of the models, significantly reducing the number of epochs compared to the minimum norm method and obtaining the best performance among the evaluated methods.

\subsection{Handwritten Digit Problem}

\begin{figure}[b]
    \centering
    \includegraphics[width=0.9\textwidth]{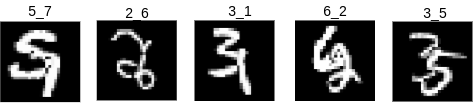}
    \caption{Images examples from the MNIST dataset modified for multi-task learning. Images have a size of $28\times28$ pixels.}
    \label{fig:mnist}
\end{figure}

In order to validate that the proposed method is independent of a specific application, we performed the experiment in the image domain considering the dataset of handwritten digits \citep{LeCun1998}. To turn it into a multi-task learning problem, we use the idea of adapting the MNIST dataset  \citep{Sabour2017,Sener2018}, in which each image is composed by the overlapping of two digit images (Figure \ref{fig:mnist}). For this purpose, two randomly selected 28x28 images were placed in the upper-left and lower-right corners of a 32x32 image. Then, the resulting image was resized to 28x28. The training and validation images were generated splitting the train set images, and the test images using the test set from MNIST dataset. A total of 50,000 training images, 10,000 validation images and 10,000 test images  were generated. 

\begin{figure}[!b]
    \centering
    \includegraphics[width=.95\textwidth]{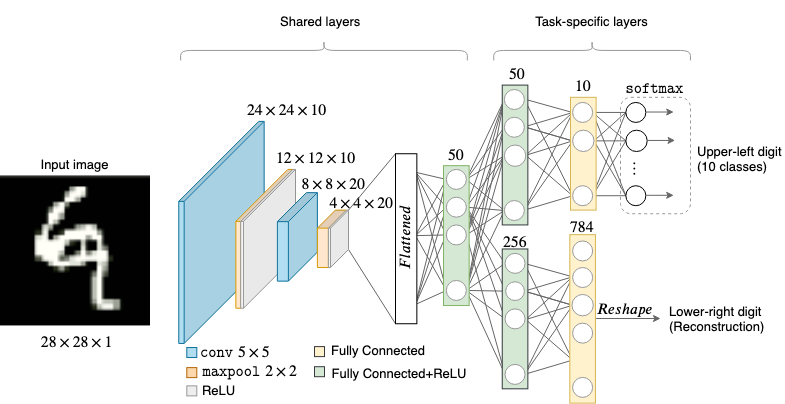}
    \caption{Multi-task learning architecture for upper-left digit classification and lower-right digit reconstruction. Best visualized with color.}
    \label{fig:lenet}
\end{figure}

In the original proposal, the problem is to classify the digits in the upper left and lower right corners. However, since all digits are from the same dataset, note that both tasks have the same nature, changing only the position where they are located in the image. In order to change the scale and nature of the problem, we defined the first task as classification of the upper-left digit and the second as an autoencoder to reconstruct the lower-right digit. The intuition behind the problem is to keep tasks with complementary information, but make them conflicting during training. Therefore, while the autoencoder will try to compress the input data into a latent space and reduce the reconstruction error, the other task will try to extract representative features to correctly classify the upper-left digit.

We use the modification of the LeNet architecture shown in Figure \ref{fig:lenet} \citep{LeCun1998,Sener2018}. The models were trained considering the Stochastic Gradient Descent (SGD) optimization algorithm with momentum and a set $lr=$ \{$1\mathrm{e}{-3}$, $5\mathrm{e}{-3}$, $1\mathrm{e}{-2}$, $5\mathrm{e}{-2}$, $1\mathrm{e}{-1}$\} of learning rates. Thus, we selected the model that obtained the best validation performance to report the results.

The Negative Log-likelihood (NLL) loss was used for the classification task, and the MSE loss for the autoencoder. Each model was trained ten times with a different seed value to get deterministic random data. Furthermore, the training process was interrupted only when there was no improvement in validation metrics during 10 epochs. 

\begin{table}[!b]
\centering
\begin{tabular}{rcc}
\hline
\multicolumn{1}{c}{\begin{tabular}[c]{@{}c@{}}Descent direction \\ method\end{tabular}} & \begin{tabular}[c]{@{}c@{}}Upper-left digit\\ Classification accuracy $\uparrow$\end{tabular} & \begin{tabular}[c]{@{}c@{}}Lower-right digit\\ Reconstruction error $\downarrow$\end{tabular} \\ \hline
Single-task                                                                             & $0.9639\pm0.0024$                                                              & $0.0181\pm0.0017$                                                                 \\
Uniform weighting                                                                                 & $\mathbf{0.9714\pm0.0018}$                                                            & $0.0418\pm0.0021$                                                                \\
Min-norm                                                                                   & $0.7393\pm0.0213$                                                              & $0.0552\pm0.0042$                                                                \\
Central dir.                                                                                & $0.8473\pm0.0107$                                                              & $0.0350\pm0.0010$                                                                \\
Min-norm + tensor                                                                          & $0.9627\pm0.0015$                                                              & $\mathbf{0.0181\pm0.0005}$                                                                \\
Central dir. + tensor                                                                       & $0.9649\pm0.0015$                                                              & $0.0185\pm0.0003$             \\ \hline                                                  
\end{tabular}
\caption{Average results of the digit classification and reconstruction tasks in the test set. We report the accuracy for classification (higher is better) and MSE for reconstruction task (lower is better).}
\label{tab:mnist}
\end{table}

Table \ref{tab:mnist} shows the average performance of the models evaluated in the test set. The first row represents the accuracy and reconstruction error of the models separately trained for each task. Therefore, we use them as a baseline to evaluate the multi-task models. As presented in the ablation studies, the common descent direction obtained by the uniform weighting tends to give greater importance to the task with the highest gradient norm. In this experiment, uniform weighting was the method that resulted in the best accuracy for the classification problem (second row). However, in addition to extracting relevant features to classification, the shared encoder must be able to compress the input data (${\rm I\!R}^{784}$) into a latent space representation (${\rm I\!R}^{50}$), so that the task-specific decoder can then reconstruct this representation. In this case, since the shared encoder was mainly optimized in relation to the classification problem, the uniform weighting reconstruction error was greater than the individual task.

The central direction method (fourth row) obtained the result a little better than the minimum-norm method (third row). However, both also had a lower performance than separately training each model. 

\begin{figure}[t]
    \centering
    \includegraphics[width=.7\textwidth]{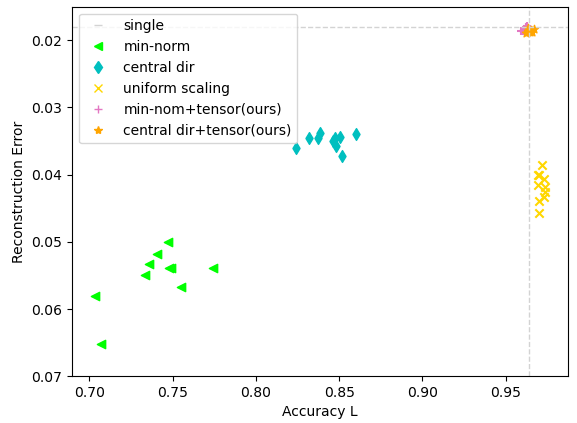}
    \caption{Plot of all trained models with the average performance of single tasks as baseline. 
    The proposed method resulted in a good balance between tasks, resulting in a performance similar or even better than single-task models. Upper-right is better. Best visualized with color.}
    \label{fig:mnist_acc}
\end{figure}

Finally, the last two rows present the use of tensors to modify the minimum-norm and central directions. Although the proposed method did not achieve the best performance separately for each task, it resulted in the best balance among the evaluated methods. This highlights that there are many trade-offs between conflicting tasks, and the proposed method was able to find a balance that maximizes the performance of all of them. Thus, while we maintain a similar performance to those models trained dedicatedly for each task, we also have the advantage of reducing the use of computational resources and inference latency, since most of the model parameters are shared between tasks. This can be better observed in Figure \ref{fig:mnist_acc}, which shows the performance of the ten trained models of each method.

To simplify the visualization, we inverted the $y$-axis of the plot. Thus, the methods with the best results are those closest to the upper-right corner.

\section{Conclusion}\label{sec:conclusion}

In this work, we address the simultaneous learning of conflicting tasks that have differences in the magnitude scale. Our method is motivated by the difficulty in performing simultaneous learning of multiple tasks in a hard parameter sharing approach without harming the performance of tasks. To this end, we propose to dynamically change the common descent direction between multiple tasks considering their convergence behavior, ensuring that the new direction will be feasible for all of them.

Experimental results showed that the proposed method outperforms the state-of-the-art methods in  simultaneous learning of conflicting tasks. Thanks to the use of temporal information to properly adjust the importance of each task, it was possible to maximize performance without harming any task.
Furthermore, the proposed method provided consistent performance with low variance in the tested problems.

Since our method is independent of a specific task, it is possible to extend our approach to other applications, such as robotics or intelligent vehicles, where several related tasks must be estimated simultaneously with reduced inference time and computational resource usage.
In addition to multi-task learning, we believe that it is also possible to use the proposed method to explore other areas where it is necessary to find a trade-off between learning multiple tasks, such as to deal with the problem of catastrophic forgetting in continuous learning, restricting the learning of a new task to be according to a descent direction that is representative for the previously learned tasks.

\hfill\break
\noindent\textbf{Acknowledgement:} The authors would like to express their gratitude to Luiz Ricardo Takeshi Horita for insightful discussions and feedback.

\hfill\break
\noindent\textbf{Funding:} This work was supported by the S\~{a}o Paulo Research Foundation - FAPESP [grant number 2019/03366-2] and the Coordination of Improvement of Higher Education Personnel - Brazil - CAPES [Finance Code 001]. This work was also supported in part by the Brazilian National Council for Scientific and Technological Development - CNPq [grant number 465755/2014- 3], by CAPES [grant number 88887.136349/2017-00], and the FAPESP [grant number 2014/50851-0].

\bibliographystyle{elsarticle-num}
\bibliography{references}

\end{document}